\documentclass[letterpaper, 10 pt, conference]{ieeeconf}  % Comment this line out if you need a4paper

\IEEEoverridecommandlockouts                              % This command is only needed if 
\usepackage{amsmath,amsfonts}

\usepackage{algorithm}
\usepackage{algpseudocode}
\usepackage{array}
\usepackage[caption=false,font=normalsize,labelfont=sf,textfont=sf]{subfig}
\usepackage{textcomp}
\usepackage{stfloats}
\usepackage{url}
\usepackage{verbatim}
\usepackage{cite}
\usepackage{bm}
\usepackage{multirow}

\usepackage{wrapfig}
\usepackage{siunitx}
\usepackage{graphicx}
\usepackage{subfig}
\usepackage{hyperref}

\title{\LARGE \bf
Task-Driven Computational Framework for Simultaneously Optimizing Design and Mounted Pose of Modular Reconfigurable Manipulators
}

\author{Maolin Lei$^{1,2}$, Edoardo Romiti$^{1}$, Arturo Laurenzi$^{1}$, and Nikos G. Tsagarakis$^{1}$% <-this % stops a space
\thanks{$^{1}$Humanoids and Human Centered Mechatronics Research Line, Istituto Italiano Di Tecnologia (IIT) Genoa, Italy. 
        {\tt\small name.surname@iit.it}}%
\thanks{$^{2}$Department of Informatics, Bioengineering, Robotics, and Systems Engineering (DIBRIS), University of Genova, Genoa, Italy
        }%
}

\begin{document}

\maketitle
\thispagestyle{empty}
\pagestyle{empty}

\begin{abstract}
Modular reconfigurable manipulators enable quick adaptation and versatility  to address different application environments and tailor to the specific requirements of the tasks. Task performance significantly depends on the manipulator's mounted pose and morphology design, therefore posing the need of methodologies for selecting suitable modular robot configurations and mounted pose that can address the specific task requirements and required performance. Morphological changes in modular robots can be derived through a discrete optimization process that involves the selective addition or removal of modules. In contrast, the adjustment of the mounted pose operates within a continuous space, allowing for smooth and precise alterations in both orientation and position. This work introduces a computational framework that simultaneously optimizes the pose and morphology mounted on modular manipulators. The core of the work is that we design a mapping function that \textit{implicitly} captures the morphological state of manipulators in the continuous space. This transformation function unifies the optimization of mounted pose and morphology within a continuous space. Furthermore, our optimization framework incorporates a array of performance metrics, such as minimum joint effort and maximum manipulability, and considerations for trajectory execution error and physical and collision constraints. To highlight our method's benefits, we compare it with previous methods that framed such problems as a combinatorial optimization problem and demonstrate its practicality in selecting the modular robot configuration for executing a drilling task with the CONCERT modular robotic platform.
\end{abstract}

\section{Introduction}

%Manipulators in flexible manufacturing and automation are designed for adaptability. 
Modular manipulators are an intuitive solution to the increasing need for mass-customized manipulation tasks, {\color{black} such as pick-and-place operations in production lines, drilling in construction environments and peg-in-hole assembly, which are common across various industries.} Specifically, these manipulators, composed of multiple interchangeable body modules, enable rapid and reversible assembly in various morphologies, that is, the form and structure of a modular robot can be tailored for specific tasks, with assembly in a few minutes~\cite{romiti2021toward}. 
%Additionally, the modular robot's mounted platform can be positioned anywhere in the environment, %allowing for flexible mounting of the robot's different morphologies. Consequently, reconfigurable %modular manipulators provide the flexibility and versatility to create task-specific kinematic %structures and mounted pose required for various applications.
Determining the morphology of a modular manipulator involves selecting and arranging various modules to assemble a kinematic structure tailored to the needs of particular tasks. This process requires a traversal through modular combinatorial spaces to identify viable morphologies. However, given the exponential increase in possible permutations from adding new modules, conducting an exhaustive search for the various morphologies becomes impractical. Moreover, for different morphologies, changing the mounted pose of the manipulator affects both the end effector's precision and performance in executing the specified task~\cite{cursi2022optimization,qin2022install,du2024learning}.

Due to the significant computational effort required to evaluate the vast discrete morphological design space for candidate designs, heuristic-based search methods ~\cite{chung1997task,romiti2023optimization,kulz2023optimizing,chen1998automatic,zhao2020robogrammar} have been effectively adapted for the optimization of modular robot design, streamlining the process by efficiently condensing the combinatorial design space to identify high-performing designs.
Meanwhile, optimizing modular robotic manipulators also involves the simultaneous adjustment of the robot's mounted pose due to its significant impact on the morphological configuration performance. The complexity of the optimization problem escalates when it includes the simultaneous optimization of the robot's morphology within a discontinuous design space and its mounted pose within a continuous Cartesian space, presenting the challenge of navigating two fundamentally different optimization spaces concurrently. The previous study in~\cite{romiti2023optimization} addresses this challenge by discretizing the mounted pose into various states, thus converting the optimization of the mounted pose into a selection process. This method still approaches the simultaneous optimization of mounted pose and morphology as a combinatorial optimization problem. 

In this work, we introduce a computational framework for simultaneously optimizing the physical morphology and the mounted pose of modular manipulators, ensuring the optimized result is specifically fit for the given tasks. 
Unlike the approach in the previous work ~\cite{romiti2023optimization}, our method incorporates the development of a mapping function that \textit{implicitly} represents the manipulator's morphology, facilitating the transition of the manipulator's discontinuous morphological states into a continuous domain. We then apply the sample-efficient Covariance Matrix Adaptation Evolutionary Strategy (CMA-ES)~\cite{hansen2003reducing} to efficiently search for high-performance solutions within this new framework. Importantly, the mapping function draws inspiration from Natural Language Processing (NLP), since we transform the robot discrete morphology state into a continuous spectrum by assigning variably continuous state variables to each module, a concept parallel but not identical to NLP's context-driven word selection probabilities~\cite{radford2018improving,yu2022interaction}.

\begin{figure}[h]
\vspace{0.35cm}
  \centering
\includegraphics[width=0.48\textwidth]{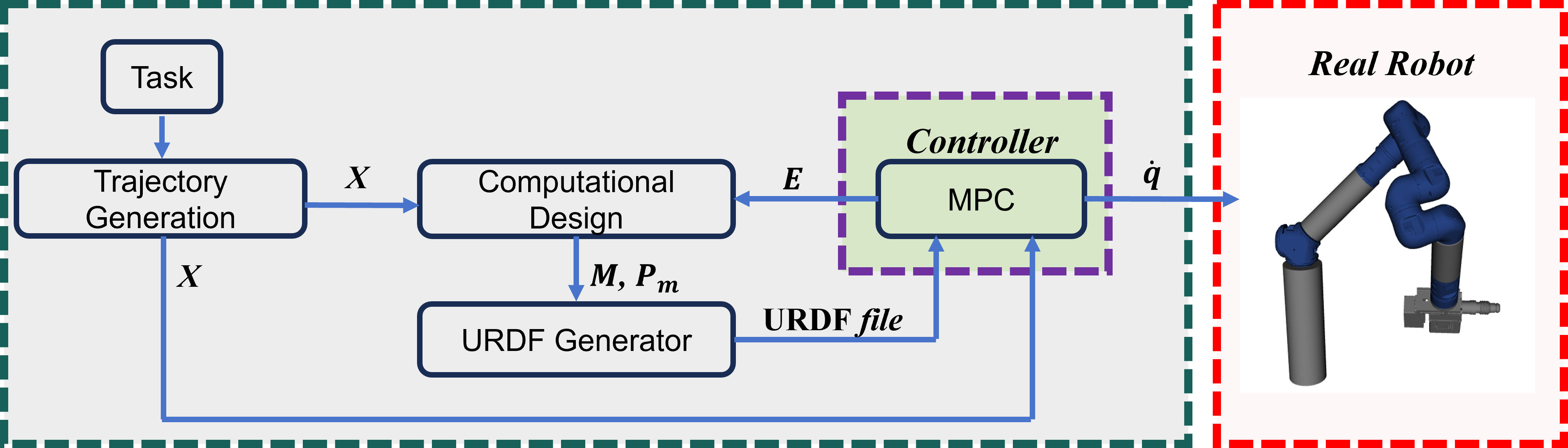}
 % \vspace{-0.25cm}
  \caption{Schematic for the computational framework}
\vspace{-0.3cm}
  \label{framework}

\end{figure}

Additionally, we integrate application-specific constraints, such as dynamic boundaries and collision avoidance, to ensure the manipulator's practical applicability. In addition, joint space optimization objectives are also considered to refine optimal results for varying demands of different application scenarios. In particular in field environments where battery conservation is critical, selecting a robot morphology that minimizes \textit{effort} is advantageous. On the other hand, in situations where energy consumption is not a primary concern, opting for a morphology that maximizes \textit{manipulability} may be more beneficial, especially to address challenges related to singularity. 
In this framework, manipulation tasks were defined as trajectories in Cartesian space, consisting of a series of desired end-effector poses. We utilize Model Predictive Control (MPC)~\cite{lei2022mpc} to control the robot to execute the specified trajectory. The execution's performance, assessed through designated evaluation metrics, is then leveraged to evaluate different morphologies and mounted pose. Additionally, MPC is able to be applied to control the actual robot in real experiments across varied morphologies. The diagram of our computational framework is shown in Fig.\ref{framework}.
The main contributions of this work include:
% (i) The mapping function was designed to transform the optimization variables for the manipulator's morphology and mounted pose both into continuous ones, enabling their simultaneous optimization within a continuous space. We compare it with genetic algorithms(GAs) to show the advantage of our method, which is commonly used to address combinatorial optimization problems.

(i) {\color{black}The mapping function was designed to transform the optimization variables for the manipulator's morphology and mounted pose into continuous variables, allowing for their simultaneous optimization within a continuous space. To demonstrate the advantages of our method, we compared it with genetic algorithms (GAs) which is commonly used to tackle combinatorial optimization problems.}

(ii) A two-level optimization framework was proposed, first ensuring task feasibility through considerations of task executing accuracy, collision avoidance and dynamic boundaries. The framework then optimizes performance by maximizing manipulability or minimizing joint effort, further refining the optimal results. We analyze the optimal results by considering various objective functions to demonstrate the effectiveness of our method. 

(iii) In our real-world experiments, the CONCERT mobile modular manipulator equipped with a 10kg drilling end-effector continuously is able to execute drilling tasks at various locations. Despite the heavy payload in executing such tasks and two drilling points close to the base link's horizontal level, the computational results still ensure collision-free with the mounted platform and satisfy dynamic constraints of the robot.

\section{Related Work}
% \subsection{Computational design for modular robot}

Since modular manipulator design spaces are discontinuous, most related research focuses on addressing morphology optimization as a combinatorial optimization problem. This approach requires a comprehensive exploration of the design space to identify the optimal morphology of the modular manipulator.
One effective exploration method is to enumerate all possible morphologies of the modular manipulator, as discussed in studies by~\cite{liu2020optimizing, romiti2021minimum,kulz2023optimizing,sathuluri2023robust}. This method ensures comprehensive coverage in the search process. However, the enumeration approach demands significant computational resources due to the large number of potential morphologies that need to be explored. To reduce this computational load, the studies in~\cite{liu2020optimizing, romiti2021minimum,kulz2023optimizing,sathuluri2023robust} introduced incidence metrics to simply the enumeration's search space, making it more efficient for practical application. However, this improved efficiency might result in less comprehensive design space exploration.

Another method utilizes population-based stochastic optimization algorithms to address the combinatorial optimization problem in exploring the morphology of modular manipulators.
Specifically, genetic algorithms~\cite{chung1997task,chen1998automatic,romiti2023optimization,kulz2023optimizing} are used to maintain and update a population of potential solutions. This approach includes processes such as mutation, crossover, evaluation, and selection, all aimed at identifying high-quality candidate individuals within the population. Compared to the enumeration approach, these methods can be more efficient. However, they are sensitive to parameters like population size, mutation rates, and crossover rates. Additionally, there is limited evidence to demonstrate their convergence to global or even local minima~\cite{sivanandam2008genetic}.
A further method for optimizing the modular manipulators' morphology employs heuristic-based search strategies. This approach requires mapping the design space to a search graph and performing evolutionary searches within this search graph~\cite{ha2018computational,zhao2020robogrammar,koike2023simultaneous}. The main challenge for the heuristic-based search method lies in finding an effective mapping function and developing a practical heuristic function to guide the search. The work in~\cite{ha2018computational} introduced the approach inspired by the A* algorithm, which formulated the design optimization as a shortest path-finding problem with a predefined heuristics function. Meanwhile, the work~\cite{zhao2020robogrammar,koike2023simultaneous} proposed to iteratively learn heuristic functions from evaluated design's results with different given tasks. The latter method simplifies the heuristic design process and demonstrates the potential for broader application of modular robots in various environments.

The design automation methods previously discussed are typically addressed using approaches related to combinatorial optimization due to the non-continuous nature of the design space. However, recent work by~\cite{hu2023glso} proposed to employ the Variational Autoencoder (VAE) network to map the discontinuous design space into a continuous latent space. This transformation allows the design optimization problem to be viewed as a continuous black-box optimization problem, potentially enhancing search efficiency compared to using related methods in combinatorial optimization. However, VAE training presents a significant challenge due to the need for high-quality data sets. The approach described in~\cite{hu2023glso} involves collecting data based on a pre-trained model. However, ensuring the quality of the data set remains challenging without the pre-trained models~\cite{hu2023glso}.

\section{Preliminaries}
\subsection{Hardware Description}
This subsection briefly delineates the hardware components for modular reconfigurable manipulators. The modular manipulator system consists of several pre-designed body modules; each module can be interchangeably connected with another. These body modules are classified into two main categories: joint and passive.

\textit{(i)} Joint Modules: The joint modules are categorized into two distinct types: "straight" and "elbow." The straight-type joint modules are actuated along the common central normal to the modular input and output flange interfaces. Conversely, the elbow-type joint modules pivot around a central axis perpendicular to the common central normal. 
\textit{(ii)} Passive Link Modules: In contrast to their active counterparts, passive modules are devoid of internal motors and are available in both straight and elbow configurations. The straight-type passive modules are further differentiated by their lengths, including 0.3m, 0.4m, and 0.6m passive link modules.

\begin{figure*}[h]
\vspace{0.2cm}
  \centering
\includegraphics[width=0.89\textwidth]
{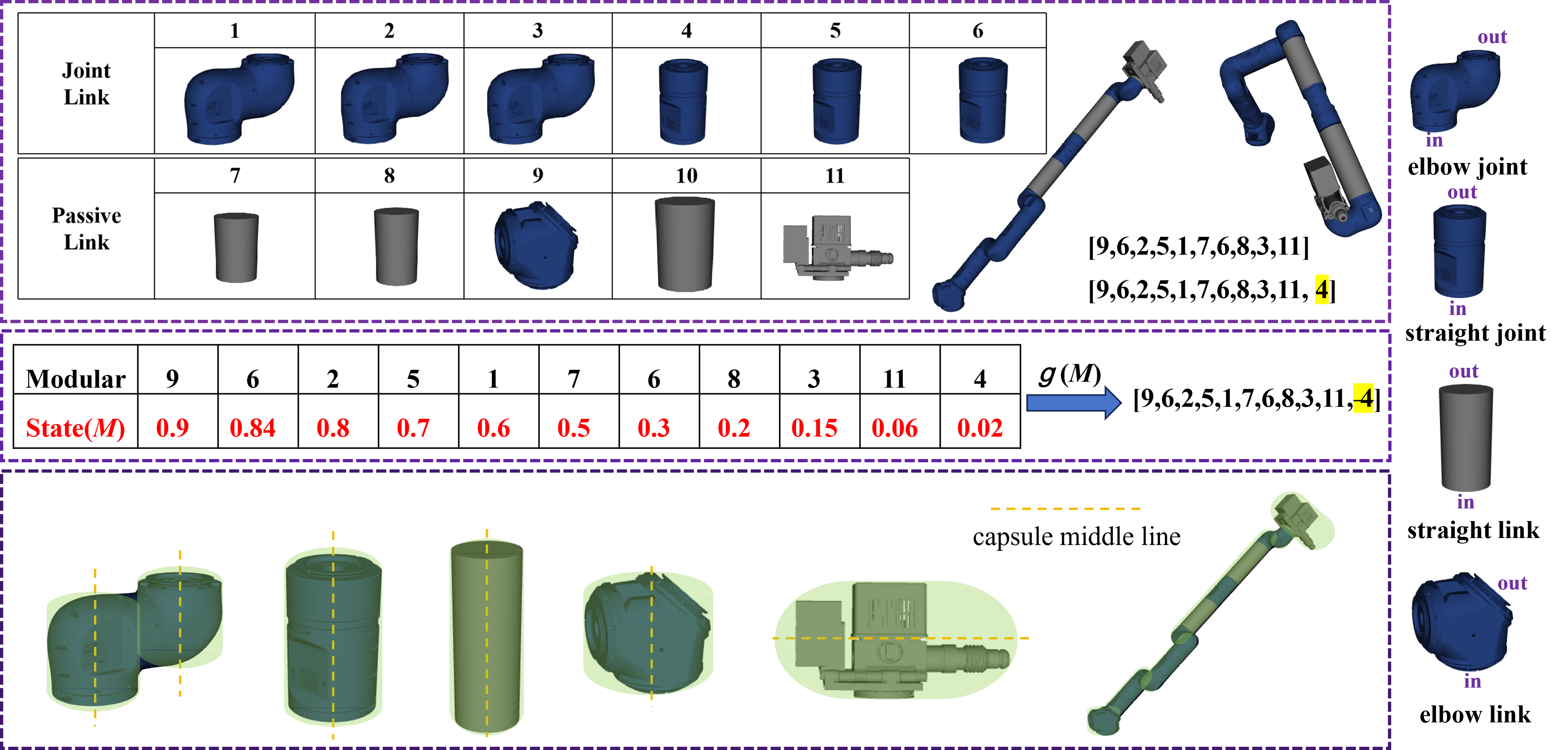}
  \vspace{-0.25cm}
     \caption{Top left: Encoding of modules by numbers, illustrated with a morphology state example. Middle left: An example of mapping modular states to morphology, where the red word indicates the state variable for different modules. Among them, the 9th module's state variable is the highest, followed by the 2nd module's state variable as the second highest. Bottom left: Representation of the collision model. Right: Types of modules with "in" indicating connection from the previous module and "out" showing the connection channel for subsequent modules. }
  \label{encoder}
  \vspace{-0.5cm}
\end{figure*}

The modular manipulator is mounted on a mobile platform, which can move to any desired pose within the given environment. For detailed information on the hardware components and the types of modules, readers can refer to~\cite{romiti2021toward} for details on our first-generation modular robot\footnote{\href{https://alberobotics.it/}{https://alberobotics.it/}}, and to Fig.~\ref{encoder}(right) for specifics.

\subsection{Controller}
The task to be executed by the modular manipulator is defined as the manipulator end-effector's desired trajectory.
For this purpose, we define the internal model state \(\bm{X}\) as follows:
\(
\bm{X} := [\bm{p}\ \ \bm{o} \ \ \bm{\dot{p}} \ \ \bm{\omega} ]^T \in \mathbb{R}^{13},
\)
where \(\bm{p} \in \mathbb{R}^{3}\) and \(\bm{\dot{p}} \in \mathbb{R}^{3}\) denote the position and linear velocity of the robot's end-effector in the world frame \(\{\bm{W}\}\), respectively. The quaternion \(\bm{o}=[\eta, \bm{\epsilon}]\in \mathbb{R}^{4}\) represents the end-effector's orientation relative to \(\{\bm{W}\}\), with \(\|\bm{o}\|=1\), where \(\eta=o_w\) and \(\bm{\epsilon}=[o_x,o_y,o_z]^T\). \(\bm{\omega} \in \mathbb{R}^{3}\) represents the end-effector's angular velocity relative to  \(\{\bm{W}\}\).
We then employ MPC as the controller for the modular manipulator. 
Using the kinematic model, we relate joint velocities \(\bm{\dot{q}}\) to the end-effector velocity \(\bm{v}_e\) via the Jacobian matrix \(\bm{J}\), expressed as \(\bm{J\dot{q}} = \bm{v}_e\). The end-effector's acceleration is calculated using \(\bm{\dot{J}\dot{q}} + \bm{J\ddot{q}} = \bm{\dot{v}}_e\). According to the kinematics of the manipulator, the discretized state-space equation, following the forward Euler integration with step size $dt$, is:
\begin{equation}
\bm{X_{k+1}}=
\left[ \begin{array}{c}
 \bm{p_k} + \bm{\dot{p}_{k}}dt \\ 
  \bm{o_k} +  \bm{o_k} \otimes [0,\bm{\omega_k}/2 ]^T dt \\ 
\bm{v_{k}} + \bm{\dot{J}_k\dot{q}_k} dt  + \bm{J_k}(\bm{\dot{q}_k} - \bm{\dot{q}_{k-1}})  \\ 
\bm{\omega_{k}} + \bm{\dot{J}_k\dot{q}_k} dt  + \bm{J_k}(\bm{\dot{q}_k} - \bm{\dot{q}_{k-1}})
\end{array} 
\right ]
\label{Dis_stat_space_detail_Old}
\end{equation}
where $k$ represents the node index of the MPC's prediction horizon.
The MPC ensures the task execution within physical limits:
\begin{equation}
\begin{aligned}
\min_{\bm{u}} \sum_{k=1}^{N_h} \bigg(  & \lVert \bm{X}(k+1) - \bm{X}_{\text{ref}}(k+1) \rVert_{{\textbf{Q}}_k} + \lVert \bm{u}(k) \rVert_{{\textbf{P}}_k} \bigg) \\
% &\text{s.t}: (\ref{eq:Dis_stat_space_detail_Old})\\
&\text{state-space equation} \\
&\text{joint constraints} 
\label{optimal_based_first_MPC}
\end{aligned}
\end{equation}
The solve results \(\bm{u}\) can be used for controlling the real robot. For more details, readers can refer to~\cite {lei2022mpc}.

\section{Morphology and Mounted Pose Co-Optimization}
\subsection{Manipulator Morphology Representation}

The selection and assembly of various modules in various sequences leads to different final morphologies for the robot. To accurately represent the morphology of the modular manipulator, we devised an encoder method for the various modules. Assuming our system comprises \( m \) modules, each module is assigned a unique numerical identifier. These identifiers are assigned from 1 to \( m \), where each number uniquely represents a module. The identified end-effector number is also designated as \( m+1 \). In cases where multiple units of the same module type are present, each unit is assigned a unique numerical code to differentiate it, even though they are the same. 
This approach of unique identification eliminates the need to impose numerical constraints on module quantities in subsequent optimization problems. 
Since the modular manipulator is structured in series, the morphology of the manipulator can be represented as a vector \( \bm{C}_v \), known as the morphology state, where \( n \) represents the dimension of the vector and the total number of modules used. 
To clarify the relationship between the actual morphology of the robot and the represented morphology state vector \( \bm{C}_v \), we provide the following details:

\textit{(i)}: In the morphology vector \( \bm{C}_v \), the value at each position directly corresponds to the module assembled in that order. Specifically, the first element in this vector represents the first module in the assembly sequence, the second element represents the second module, and so on. 

\textit{(ii)}: The total number of the modules $m+1$ can be predetermined. This information allows us to define the maximum dimension of the morphology vector \( \bm{C}_v \in \mathbb{R}^{n} \). The number of modules used in the manipulator should be less than or equal to \( m+1 \)(the total number of available modules), i.e., \( n \leq m+1 \).

\textit{(iii)}: The concept of the \textit{End of Manipulator} (EoM) is introduced, represented by the m+1 module, and serves as the end-effector of the manipulator. The EoM module in the sequence also signifies the completion of the assembly. Therefore, any modules after the EoM module are not considered part of the connected manipulator's morphology.

This representation method is similar to sentence representation in NLP, wherein the concept of 
% \textit{End Of Manipulator} 
EoM parallels the  \textit{End Of Sentence} (EoS)~\cite{radford2018improving} marker in NLP. In NLP, words following the EoS are not analyzed, and similarly, in the proposed modular robot representation, modules listed after the EoM are not to be assembled. 
To further illustrate this method with an example, assuming that there are 10 distinct modules (\( m=10 \)), the end-effector module is numbered as \( m+1=11 \). In this case, the modular manipulator with a state of morphology represented by the vector \([9,6,2,5,1,7,6,8,3,11, 4]\) is effectively equivalent to the state of morphology represented by \([9,6,2,5,1,7,6,8,3,11]\),
since any modules listed after the 11th EoM in the sequence are not considered for assembly with the preceding modules. 
In other words, if the framework generates two distinct morphological states, such as \([9,6,2,5,1,7,6,8,3,11,4]\) and \([9,6,2,5,1,7,6,8,3,11]\), despite the differences in their lengths, module \(4\)—positioned after the EoM, module \(11\)—will not be considered for inclusion as one of the modules in the final manipulator morphology. The example of the manipulator's morphology state is shown in Fig.~\ref{encoder}.

%If all available modules are used in the assembly, such that the dimension of the morphology vector equals 10, then the assembly is still deemed complete, regardless of whether the EoM (11th module) is explicitly present. If the EoM is not present, even though the vector length reaches or is less than the predetermined maximum of 11, the assembly is still considered complete. For example, a morphology represented by the vector \([9,2,5,1,7,6,8,3,11]\) is same as one represented by \([9,2,5,1,7,6,8,3]\). This method provides a flexible way to represent different possible morphologies of the manipulator while also considering the manipulator's length. The example of the manipulator's morphology state is illustrated in Fig.~\ref{encoder}.

\subsection{Mapping Function for Morphology Representation} 

We introduce a mapping function \( g(\bm{.}) \), designed to transform this discrete morphology state into a continuous module state \( \bm{M}  \in R^{m+1}\). 
%
% The mapping function \( g(\bm{.}) \) in our computational framework draws an intriguing parallel to the concept of word generation probabilities in NLP. In NLP, each word in a sentence has a certain probability of being generated or selected, influenced by the context set by preceding words. This probabilistic approach underscores that word selection is not merely a discrete choice but is guided by continuously varying probabilities~\cite{radford2018improving,yu2022interaction}. Similarly, Our framework assigns each modular component a state variable that varies continuously. 
% This way, 
The dimension of \( \bm{M} \in \mathbb{R}^{m+1} \) is fixed, equaling the total number of modules (\( m+1 \)), where \( m+1 \) includes all the body modules  and the end-effector of the modular. The first element in \( \bm{M} \) represents the state of the module with the encoded number \( 1 \), the second element corresponds to the module with the encoded number \( 2 \), and so on. By implementing the mapping function as \( \bm{C}_v = g(\bm{M}) \), the encoded results of each module's state in \( \bm{M} \) implicitly represent the robot's morphology state \( \bm{C}_v \). An example is illustrated in the middle of Fig.~\ref{encoder}, where each module exhibits different states. These states can \textit{implicitly} represent the morphology of the manipulator according to the mapping function.

With different states of various modules, the mapping function \( g(\bm{.}) \) selects and assembles modules in descending order based on their state variable values. 
It should be noted that if the given states of modules are the same, then a random method will be employed to sort the modules with the same state, ensuring that each module with the same state has an equal probability of being sorted first.
The module with the highest value is placed first, continuing until the module with the lowest value is last. This order forms the vector of the morphology state \( \bm{C}_v \). 
% This function enables the selection and arrangement of modules according to their state variable values, mirroring the probabilistic selection of words in NLP to form coherent sentences. Thus, our method evolves from a conventional discrete module selection to one influenced by a continuum of values akin to the spectrum of probabilities in NLP word selection. 
%Meanwhile, modules appearing after the EoM in this sequence are not included in the assembly.
% This also offers the benefit that if the EoM possesses a relatively higher state variable compared to all other modules, it correlates with fewer modules being utilized in the final assembly. 
%This approach ensures that a higher EoM value correlates with fewer modules used in the final assembly.

\subsection {Morphology and Mounted Pose Optimization}

The optimization objective is to concurrently optimize both the morphology state \( \bm{C}_v \) and the mounted pose state \( \bm{P}_m \) of the modular manipulator. The morphology state \( \bm{C}_v \) is inherently discontinuous, whereas the space for the mounted pose of the robot \( \bm{P}_m \) is continuous. Here, \( \bm{P}_m = [x, y, \theta] \in \mathbb{R}^3 \) represents the mounted pose, where \( x \), \( y \), and \( \theta \) denote the position on the \( x \) and \( y \) axes and the orientation in the \textit{yaw} direction in the world frame \(\{W\}\) , respectively.
The mapping function introduced in the last section transforms the robot's discontinuous morphological state into a continuous state. Consequently, the problem of optimizing the morphology state \( \bm{C}_v \) is reframed as optimizing the module state \( \bm{M} \). The mapping function ensures that both the morphology \( \bm{M} \) and the mounted pose \( \bm{P}_m \) are in a continuous space. 

The optimization focuses on refining two optimal states, including modular state \(\bm{M} \) and its mounted pose \(\bm{P}_m \). The objective is to derive these states to achieve more accurate reference trajectory tracking while considering the given optimization function. For the control policy, regardless of the varying morphologies and mounted pose, all manipulators in this study are controlled using MPC. This controller is designed to optimize the joint states of the manipulator, represented as \( \bm{q} \in \mathbb{R}^d \), to track the reference trajectory. Here, \( d \) is the DoFs of the manipulator. 
In addition to the tracking error objective, the optimization function must include two constraints: the robot must satisfy dynamic constraints and generate collision-free motion. 
\subsubsection{Dynamic Constraints}
The complete dynamics equation for the manipulator can be expressed as:
\begin{equation}
\bm{M}(\mathbf{q})\ddot{\mathbf{q}} + \mathbf{B}(\mathbf{q}, \dot{\mathbf{q}})\dot{\mathbf{q}} + \mathbf{g}(\mathbf{q}) + \mathbf{J}^T(\mathbf{q})\mathbf{F}_{ext} = \bm{\tau}
\end{equation}
In this equation: \( \bm{M}(\mathbf{q}) \) represents the mass (or inertia) matrix, which depends on the joint positions \( \mathbf{q} \);\( \ddot{\mathbf{q}} \) is the joint acceleration; \( \mathbf{B}(\mathbf{q}, \dot{\mathbf{q}}) \) is a vector function that includes Coriolis and centrifugal forces; \( \dot{\mathbf{q}} \) denotes the joint velocity; \( \mathbf{g}(\mathbf{q}) \) represents the torque or force due to gravity; \( \mathbf{\bm{\tau}} \) is the torque applied to the joints; \( \mathbf{J}^T(\mathbf{q}) \) is the transpose of the Jacobian matrix; \(\mathbf{F}_{ext} \) is the vector of external forces acting on the manipulator.
For the dynamic constraints of the manipulator, every joint of the manipulator must ensure $||\mathbf{\tau}_i|| < \mathbf{\tau}_{\text{max}}$ with $i$ represent the $i$-th joint of the manipulator.

\subsubsection{Collision-Free Constraints}
To efficiently detect collision between the manipulator with various morphologies and the environment, each module is approximated as a \textit{capsule}(Fig. \ref{encoder}). With these predefined capsule dimensions and parameters, we can assess collisions between each manipulator module and environmental obstacles, which facilitates the detection of collisions for the entire manipulator in different morphologies. Additionally, the position of each module within the environment can be determined through forward kinematics.
To calculate the distance between a capsule and an environmental obstacle, we measure the distance between the capsule and the obstacle by using the flexible collision library (FCL)~\cite{pan2012fcl}. 
The collision constraints are formulated as follows:
\begin{equation}
\text{C}(\bm{p}_r) \geq d_{\text{safe}}, \ \forall \bm{p}_r \in \text{Robot},
\end{equation}
where \( \bm{p}_r \) represents the point on the robot, and \( d_{\text{safe}} \) is the minimum allowable distance between robot and obstacles.

\subsubsection{Joint Space Optimization}

In addition to the primary objectives, we also consider an optimized state by considering the "soft objective" We design two different objective functions: one focused on minimizing \textit{effort} and the other on maximizing \textit{manipulability}. This dual-focused approach enables us to optimize for the most appropriate result based on specific operational requirements. 
For the minimizing effort function, we can write as follows:
\begin{equation}
    \text{min} \ \  F_{\text{eff}} = \frac{1}{N} \sum_{i=1}^{N} \sum_{j=1}^{d} |\tau(q_{i,j})|
\end{equation}
In this equation, \( \tau() \) represents the torque at the joint. The specific process of calculating this torque can be further referenced in \cite{romiti2021toward}.

 The second metric aims at maximizing \textit{manipulability}, which can be expressed as:
\begin{equation}
    \text{max} \ \ M_{\text{man}} =  \frac{1}{N} \sum_{i=1}^{N} {\text{det}(\bm{J}(\bm{q}_i) \bm{J}(\bm{q}_i)^T)}
\end{equation}
where \( \bm{J}(\bm{q}) \) is the Jacobian matrix of the manipulator at a given configuration \( \bm{q} \) at the $i$-th control loop, \( N \) and \( d \) represent the total number of control loops of the controller and the DoFs, respectively.

\subsubsection{Optimization Formulation}
The primary objective for the manipulator is to execute the assigned tasks proficiently while adhering to the constraints given. To ensure both the feasibility of the trajectory and the robot's safety, it is imperative that the tracking error remains below a specified threshold. In addition, the robot must satisfy dynamic constraints and avoid collisions with obstacles. 
The formulation of the optimization problem, which encapsulates these requirements, can be expressed as follows:
\begin{equation}
\begin{aligned}
    & \underset{g(\bm{M}),\textbf{P}_m}{\text{minimum}}
    & & X_e = -w e^{-w_f F_{\text{eff}} + w_m M_{\text{man}} } \\
    % & & X_e = \frac{1}{N} \sum_{i=1}^N  f(g(\bm{M}),\bm{P}_m,\bm{q}_i) \\
    & \text{subject to}
    & & \bm{q}_i \in [\bm{\underline{q}}, \ \ \bm{\overline{q}}],\\
    & & & || FK(\bm{M},\bm{P}_m,\bm{q}_i)\bm{T}_{d,i}^{-1} ||^2 \leq \xi, \ \forall i, \\
    & & & \text{C}(\bm{p}_r) \geq d_{\text{safe}}, \ \forall \bm{p}_r \in \text{Robot}, \\
    & & & \forall i \ \ ||\tau_i(\mathbf{q}_i, \dot{\mathbf{q}}_i, \ddot{\mathbf{q}}_i)|| < \mathbf{\tau}_{\text{max,i}},
\end{aligned}
\label{eq:constraint}
\end{equation}
where $w$, ${w}_f$ and ${w}_m$ represent the weight for optimization objective, optimizing minimum joint effort and maximum manipulability, respectively.
The variables $\bm{\underline{q}}$ and $\bm{\overline{q}}$ denote the lower and upper bounds of the joint state of the robot, respectively, that is, the joint limits of the manipulator.  The $FK(\bm{.})$ represents the forward kinematics results of the manipulator, with the output being the manipulator's end-effector state, $\bm{T}_{d,i}$ represents the desired state of the end-effector at $i$-th control loop, and \(\xi\) represents the threshold for the tracking error.
This final optimization result includes the modular state $\bm{M}$ and the mounted pose $\bm{P}_m$, thereby fulfilling the requirements of the task.

Then, we rewrite the optimization in a non-constrained formulation. The optimization function considers five key aspects: manipulator's manipulability, joint effort, tracking error, collision constraints penalty, and dynamic constraints penalty. %Different from the Eq.~\ref{eq:constraint}, 
The primary objective of this optimization formulation is to minimize the tracking error. If dynamic or collision constraints are not satisfied, a significant penalty is imposed on the optimization function.  Once the tracking error is diminished to below a specified threshold $\xi$, the optimization function then shifts focus to joint space for minimize joint effort or maximum manipulability.
The objective function is articulated as follows.

\begin{equation}
\begin{aligned}
    & \underset{\bm{M},\bm{P}_m}{\text{minimize}} \ \ E = \underbrace{E_{\text{track}} + E_{\text{collision}} + E_{\text{dynamic}}}_{E_{\text{sum}}} - \\ 
    & \quad \begin{cases}
        w e^{-w_f F_{\text{eff}} + w_m M_{\text{man}}} , & \text{if } E_{\text{sum}} <  \xi\\
        0, & \text{otherwise.}
    \end{cases}
\end{aligned}
\label{eq:cmaesopt}
\end{equation}
where 
\begin{equation}
    E_{\text{track}}(\bm{M},\bm{P}_m,\bm{q}) = \frac{1}{N} \sum_{i=1}^{N} || FK(\bm{M},\bm{P}_m,\bm{q}_i)\bm{T}_{d,i}^{-1} ||^2 ,
\end{equation}
\begin{equation}
    E_{\text{collision}}(\bm{M},\bm{P}_m,\bm{q}) = 
    \begin{cases} 
    \num{e10}, & \text{collision detected} \\
    0, & \text{otherwise},
    \end{cases}
\end{equation}
\begin{equation}
    E_{\text{dynamic}}(\bm{M},\bm{P}_m,\bm{q}) = 
    \begin{cases} 
    \num{e10}, & \text{dynamic constraints violated} \\
    0, & \text{otherwise}.
    \end{cases}
\end{equation}
\begin{equation}
    E_{\text{sum}} = E_{\text{track}} + E_{\text{collision}} + E_{\text{dynamic}}
\end{equation}

The function is hierarchically structured. Only when the tracking error is smaller than the given threshold, while adhering to collision and dynamic constraints, the optimization process proceed to consider further the optimization for maximum \textit{manipulability} and minimum joints' \textit{effort}. This layered approach ensures that fundamental performance metrics are prioritized before advancing to refine additional aspects of the robot's functionality. To address this optimization problem, we employ the CMA-ES as the gradient-free optimizer, considering both the optimal state's continuous space and the object's constraints. The optimal results from Eq.~\ref{eq:cmaesopt} include the modular state $\bm{M}$ and the mounted pose $\bm{P}_m$. The final morphology state can be calculated using the mapping function as \(\bm{C}_v = g(\bm{M}) \).

\section{Experiments and Evaluations}
\subsection{Experimental Setup}
For the parameters in the CMA-ES algorithm, we set the maximum number of generations to 100, and the maximum number of populations in each evaluation loop is set to 20, with sigma being 0.25. The evaluation process is executed on multi-process CPUs.
Regarding the controller parameters, the tracking gains for position and orientation in MPC are set at \(10\) and \(4\) in \(\bm{Q}_k\).
The gains for the regularizer \(\bm{P}_k\) in the controller are set at 0.0001, the horizon is set at 10, and the control frequency is 100Hz(dt = 0.01). The controller's parameters remain consistent across different morphologies of the robot. 
Meanwhile, the maximum joint torque for modules $1$, $2$, $4$, and $5$ is set at 120 Nm, whereas for modules $3$ and $6$, it is 200 Nm. The joint position limits for each module range from -2.7 to 2.7 radians and the maximum joint velocity is capped at 2.0 rad/s.
The optimization gains $w$ in Eq.~\ref{eq:cmaesopt} were set to 10, and the tracking threshold $\xi$ was set to 0.001.
Additionally, with a total of 10 different modules available, there are approximately $10! = 3,628,800$ of different morphologies that can be constructed. We have a mobile platform on which the modular manipulator can be installed. This platform can be placed in various poses within the environment.

\subsection{Task Description}
We designed a specific task to demonstrate the effectiveness of our proposed framework. This task involves using an end-effector drilling tool to drill walls, a common task in building scenarios. The task requires drilling six points on the vertical wall with the same orientation. During the drilling process, the end effector requires maintaining an orientation perpendicular to the wall, whereas orientation in the $z$-direction, relative to the end effector's framework along the axis of the drill bit, is less critical and can be overlooked. 
There are two main obstacles in the environment: the mobile platform and the walls within the environment. However, the collision problem can be ignored between the robot and the walls during the drilling process. During the drilling process, it is also necessary to consider the contact force exerted on the end-effector between the robot and the wall.
\begin{wrapfigure}{r}{0.25\textwidth} % "r" for the right side, "l" for the left side, and the width of the figure space
  \centering \includegraphics[width=0.23\textwidth]{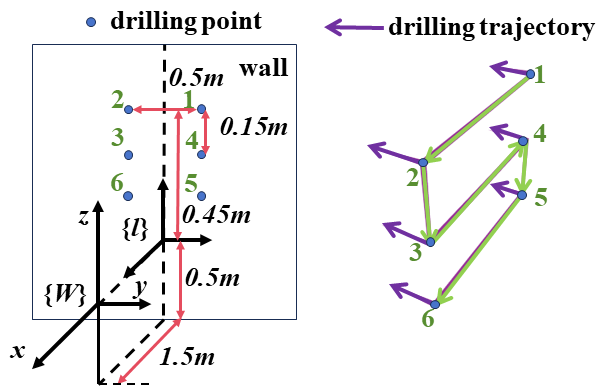}
  \caption{Arrangement of drilling points relative to the wall and world frame.}
  \vspace{-0.3cm}
  \label{fig:drillingpoints}
\end{wrapfigure}
% \begin{figure}[h]
%   \centering
% \includegraphics[width=0.35\textwidth]{figure/drillingpositiondes.png}
%   \caption{Arrangement of drilling points relative to the wall and world frame. }
%   \label{fig:drillingpoints}
%   \vspace{-0.3cm}
% \end{figure}
% \begin{wrapfigure}{r}{0.4\textwidth} % "r" for the right side, "l" for the left side, and the width of the figure space
%   \centering \includegraphics[width=0.4\textwidth]{figure/drillingpositiondes.png}
%   \caption{Arrangement of drilling points relative to the wall and world frame.}
%   \label{fig:drillingpoints}
% \end{wrapfigure}
The drilling points are specified relative to the world frame \(\{W\}\). The height between the base link and the ground is 0.5 m (the height of the mobile platform). The coordinates of the drilling points are as follows: at a distance of 1.5 m to the world frame's origin on the $x$-direction, the points alternate along the \(y\)-axis at intervals of 0.15 m forward and backward, with \(z\)-axis values descending in increments of 0.15 m from a starting height of 0.45 m. The drilling sequence corresponding locations are given by the coordinates \([-1.5, 0.15, 0.45]\), \([-1.5, -0.15, 0.45]\), \([-1.5, -0.15, 0.30]\), \([-1.5, 0.15, 0.30]\), \([-1.5, 0.15, 0.15]\), and \([-1.5, -0.15, 0.15]\) with respect to the world frame \(\{W\}\).
Fig.\ref{fig:drillingpoints} on the left illustrates the arrangement of the drilling points relative to the wall and the world frame. The dashed lines indicate the horizontal distances from the wall to the drilling points, while the solid lines with arrows represent the axes of the wall coordinate frame \(\{l\}\) and the world frame \(\{W\}\). Fig.\ref{fig:drillingpoints} on the right shows the designated 3D trajectory for the drilling task; the purple line signifies the drilling trajectory, with a depth of 0.1 meter, and the green line illustrates the trajectory to move the robot from one point to the next.

\subsection{Comparison Study}
\label{subsec:warmstart}
We conduct a comparative analysis with the baseline algorithms to demonstrate the efficiency and advantages of our proposed computational method. The weights assigned to the manipulability and effort objectives were set at 1 and 0.01, respectively, indicating that [$w_m$, $w_f$] = [1.00, 0.01]. The baseline methodology involves adapting the GA, with implementation following the approach described in~\cite{romiti2023optimization}. This approach discretizes the mounted pose into various states, effectively transforming the state of the mounted pose into a discretized format. This enables the simultaneous optimization of the robot's mounted pose and morphology while still addressing it as a combinatorial optimization problem. 
%In the context of our evaluation, we define the scenario within a work-cell, where the robot's base position is limited in \(1.2m \times 0.6m\) as same as the setting in~\cite{romiti2023optimization}. This setup spans a range from \(-0.6m\) to \(0.6m\) on the \(x\) and \(-0.3m\) to \(0.3m\) \(y\) axes, respectively, permitting a discretization step of \(0.2\) m.  Regarding rotational adjustments, we set the angle offset relative to the world frame's \(z\)-axis ranging in \( \pi \) to \(-\pi\) rad, with a discretization step of $\frac{\pi}{2}$ rad. The discretization step also mirroirs with the~\cite{romiti2023optimization}. Consequently, the mounted position yields 5 and 3 discrete states along each $x$ and $y$ axis and 4 possible states in \(z\)-axis directional orientation.
In our evaluation context, we situate the scenario within a work cell where the limitation of the robot's base position is \(1.2m \times 0.6m\), and rotationally in the world frame's \(z\)-axis, ranging from \( \pi \) to \(-\pi\), a setup that mirrors the design outlined in~\cite{romiti2023optimization}. In alignment with~\cite{romiti2023optimization}, the position discretization step was set at \(0.2m\) and the angle discretization step at \(\frac{\pi}{2}\) rad. Consequently, the mounted position established yields 5 and 3 discrete states along the \(x\) and \(y\) axes, respectively, and 4 possible rotational states along the \(z\) axis.

\begin{figure}[ht]
    \centering
    \captionsetup[subfloat]{font=scriptsize}
    \subfloat[20 population]{
    \includegraphics[width=0.22\textwidth, trim=20pt 0pt 20pt 0pt, clip]{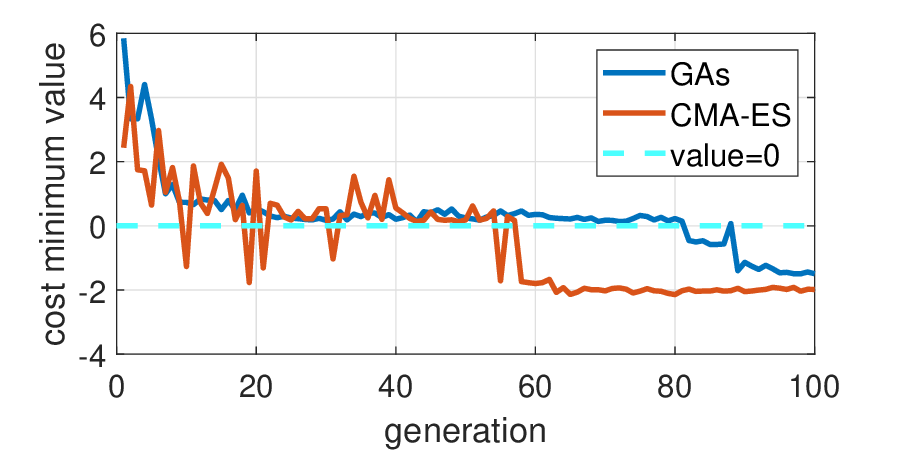}
        \label{fig:mor1_task2}
    }
    \hspace{5pt} % Adjust spacing as needed
    % Second Subfigure
    \subfloat[40 population]{
    \includegraphics[width=0.22\textwidth, trim=20pt 0pt 20pt 0pt, clip]{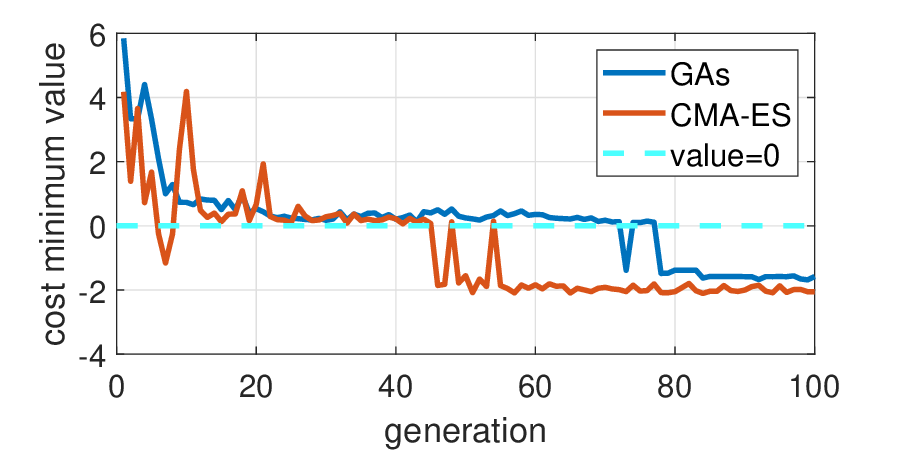}
        \label{fig:mor2_task2}
    }
    \caption{Minimum cost evolution over generations with different population. }
    \label{fig:cost}
    \vspace{-0.6cm}
\end{figure}

{\color{black} In the GA set-up, the mutation probability was set to 0.5, and the crossover probability was set to 0.1, with 100 generations. For the first comparison group, we maintained the population size at 20 for each generation, reflecting the settings used in CMA-ES. In the second comparison group, we adjusted the population size for both GAs and CMA-ES to 40, consistent with the parameters set in~\cite{romiti2023optimization}. It is important to note that, regardless of the method used (GAs or CMA-ES), the number of samples evaluated in each generation remains the same, leading to the assumption of the same evaluation time per generation.}

To execute the specified tasks, we first tackle the inverse kinematics (IK) challenge to pinpoint the trajectory's initial desired point, setting this as the MPC's initial joint state. We then simulate five distinct IK solutions as initial configurations for the joint setup to enable a productive warm start for the MPC, evaluating their performance in the process. The most efficient initial configuration state and its performance evaluation are recorded to optimize the morphology and mounted pose. These warm-start findings are also valuable for application in real-world experimental conditions.

Fig.~\ref{fig:cost} depicts the minimum value achieved by each generation when employing GAs and CMA-ES for the optimization with the different given population. It is important to note that a cost value smaller than zero indicates that the task at the first level was successfully completed. This means that the tracking error is smaller than the given threshold while also adhering to the dynamic and collision constraints. 
Compared to GAs, our computational methods demonstrate superior performance in both convergence speed and the quality of solutions for the specified cost function, regardless of population size, which is reflected in consistently lower fitness values in the figure. Notably, this improvement occurs even though the population size in the CMA-ES is 20 in every generation, which is smaller than the population size used in our previous work~\cite{romiti2023optimization}. {\color{black} Additionally, given that we assume the computation time for each generation is the same and that CMA-ES converges more quickly, we can infer that our method is computationally faster.}
%Meanwhile, although the GAs can find feasible solutions after discretizing installation positions, this method misses many optimal solutions because, in reality, the potential installation pose are infinite. 
In fact, the rationale for setting discrete parameters for comparative analysis is to align with the setting presented in the paper~\cite{romiti2023optimization}. Therefore, we adopt the parameterization outlined in this reference. In real-world applications, finer discretization of our space could be considered to enhance the quality of optimization results. However, this refinement tends to decelerate the convergence speed of GAs by broadening the spectrum of choices, thereby complicating the selection process. Moreover, note that, in practical scenarios, the potential for discretization is essentially boundless.

\begin{table}[h!]
  \centering
  \begin{tabular}{|c|c|c|c|c|c|}
    \hline
     $\textbf{\textit{M}}$ & $\textbf{\textit{P}}_m$ & $w_m$, $w_f$& $E_{\text{track}}$ & $ M_{\text{man}}$ & $F_{\text{eff}}$  \\ \hline
    C-1 & [-0.15, 0.21,  1.57] & 1, 0.01 & $\num{3.6e-5}$ & 0.41 & 198.7 \\ \cline{1-6}
    C-2  & [0.29, -0.03,  2.2] &1, 0.00 & \num{1.06e-4} & \textbf{0.83} &  329.5\\ \cline{1-6}
    C-3  & [0.13, 0.09,  1.57] & 0, 0.01 &  \num{9.1e-5} & 0.35 & \textbf{186.6} \\ \hline
  \end{tabular}
  \caption{Optimal results with different objectives}
  \vspace{-0.6cm}
  \label{optimalresult}
\end{table}

\subsection{Evaluations with Different Optimization Objectives}

\begin{figure}[ht]
  \vspace{-0.65cm}
    \centering
    \captionsetup[subfloat]{font=scriptsize}
    \subfloat[C-1]    {      \includegraphics[width=0.14\textwidth]{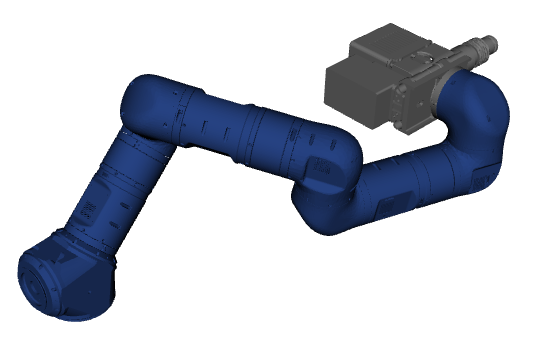}
        \label{fig:mor1_task1}
    }
    \hspace{5pt} % Adjust spacing as needed
    % Second Subfigure
    \subfloat[C-2]{
        \includegraphics[width=0.135\textwidth]{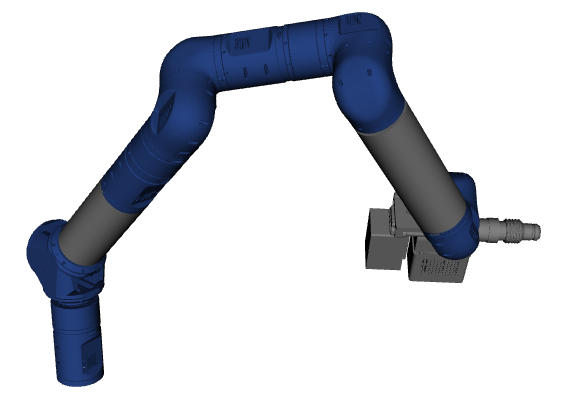}
        \label{fig:mor2_task1}
    }
    \hspace{5pt} % Adjust spacing as needed
    % Third Subfigure
    \subfloat[C-3]{
        \includegraphics[width=0.12\textwidth]{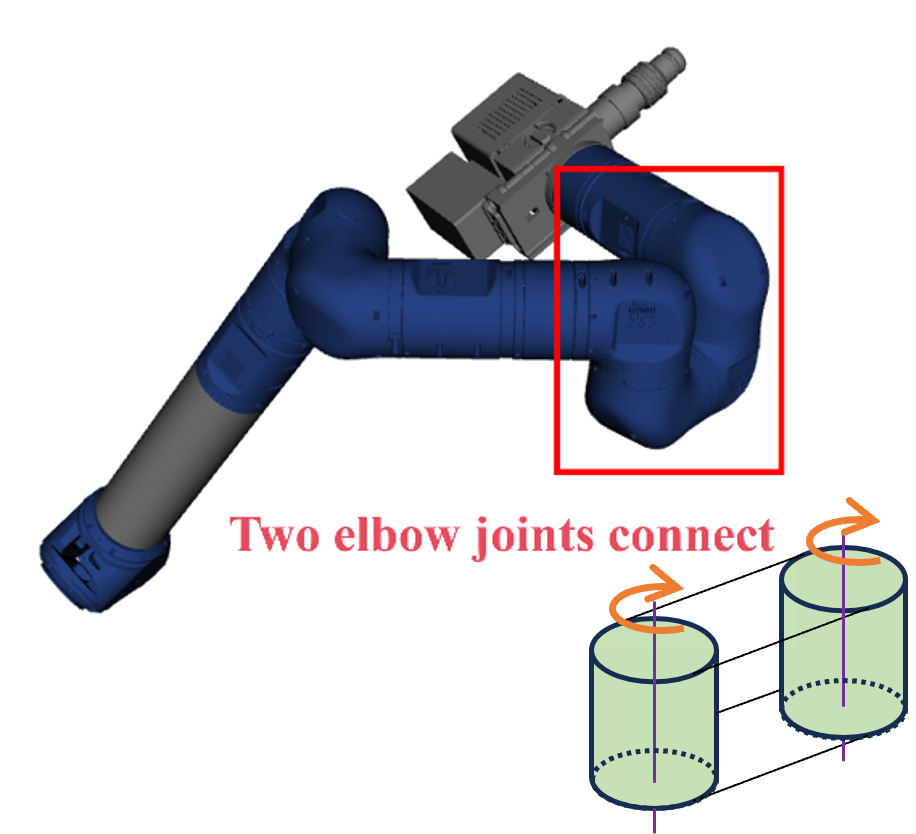}
        \label{fig:mor3_task1}
    }
    \caption{The derived manipulator morphologies for the drilling task}
    \label{fig:three_subfigures_Task1}
    \vspace{-0.25cm}
\end{figure}

We designed three different scenarios with three other optimization objectives to further showcase the effectiveness of the proposed framework. 
In the first scenario, our optimization objective combines maximizing manipulability with minimizing joint effort, with weights of 1 and 0.01 assigned to these objectives, respectively, denoted as [$w_m$, $w_f$] = [1.00, 0.01]. The second scenario maximizes the manipulability objective alone, setting its weight at 1, thus [$w_m$, $w_f$] = [1.00, 0.00]. In contrast, the third scenario focuses exclusively on minimizing joint effort, with a weight of 0.01 assigned to this objective, represented as [$w_m$, $w_f$] = [0.00, 0.01]. Given these varied optimization objectives, our computational framework identified different optimization solutions to execute the given task accordingly.
Detailed information on these specific weight assignments and optimization results is presented in TAB~\ref{optimalresult} and Fig.~\ref{fig:three_subfigures_Task1}. The $\bm{P}_m$ was represented in the world frame \(\{W\}\). In Fig.~\ref{fig:three_subfigures_Task1}, C-1 denotes the morphology optimized for the combined objective of maximizing manipulability
%(a metric indicating the robot's distance from singularity configurations) 
and minimizing joint effort, while C-2 and C-3 represent morphology optimized for maximum manipulability and minimum joint effort, respectively.
These results demonstrate that the tracking error remains below a specified threshold, while avoiding collisions and adhering to dynamic constraints. 

In the optimized results C-2, shown in TAB.~\ref{optimalresult} Second line, focusing solely on maximizing manipulability led to approximately increases by $\frac{0.83-0.41}{0.41} \approx 102.4\%$ (from $0.41$ to $0.83$) and $\frac{0.83-0.35}{0.35}  \approx 137.1\%$ (from $0.35$ to $0.83$) compared to results in C-1 and C-3, respectively.
Meanwhile, we observed that the modular robot's extended length implies the necessity for more modules to construct the manipulator. However, when comparing the morphology in C-1 and C-3, there is a noticeable increment in the joint effort to execute the given task. 
In the optimized results of C-3, presented in the third line of the TAB~\ref{optimalresult}, with the sole objective of minimizing joint effort, we observed a reduction compared to the results in C-1 and C-2. The joint effort decreased from $198.7$ to $186.6$, equivalent to a reduction of approximately $\frac{186.6-198.7}{186.6} \approx -6.5 \%$ (from C-1 to C-3), and from $329.5$ to $186.6$, which decrease approximately $\frac{186.6-329.5}{329.5} \approx -43.4\%$ (from C-2 to C-3). 
Meanwhile, we observe that parallel alignment of the manipulator's second-to-last and third-to-last rotational axes in the final optimal results leads to lower torque requirements. This reduction is due to the shared load across the parallel axes, which decreases the torque arm length — the perpendicular distance from the axis to the force's line of action. Consequently, a shorter torque arm demands less torque to generate the required rotational force, thus minimizing joint effort. However, this morphology leads to decreased manipulability; despite the robot having six degrees of freedom, the parallel alignment of two rotational axes essentially subtracts one effective degree of freedom, constraining its ability to achieve certain positions and orientations at the end-effector.

\subsection{Experiments}
\begin{figure}[h]
  \centering
\includegraphics[width=0.48\textwidth]{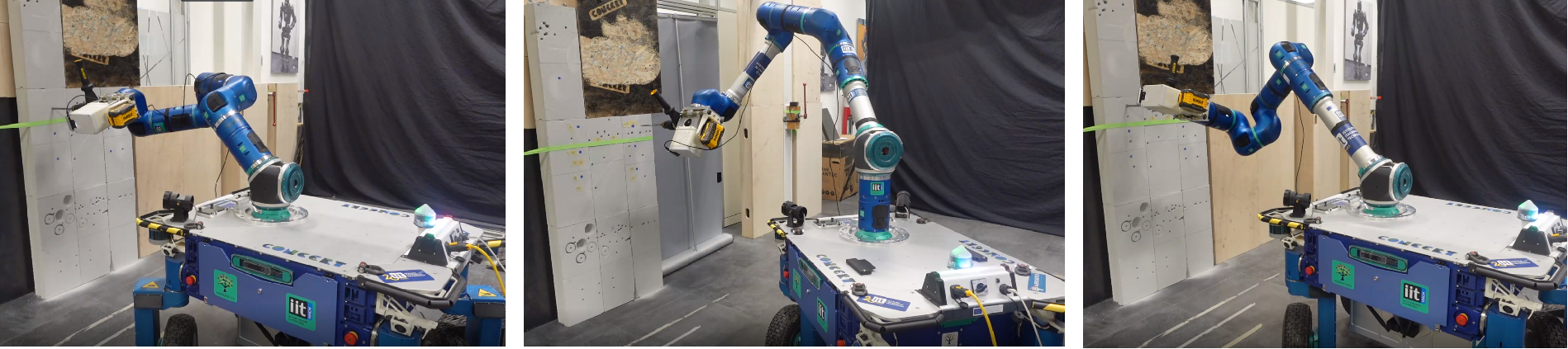}
  \caption{Snapshots of the experiments for various computational results.}
  \label{fig:experiments}
  \vspace{-0.5cm}
\end{figure}

We also performed experiments with the CONCERT modular mobile manipulation robot executing the given drilling task to further validate the optimization results. The experiments snapshots can be obtained in Fig~\ref{fig:experiments}, from left to right in the images represent the morphology from C-1, C-2, and C-3, as illustrated in Fig.~\ref{fig:three_subfigures_Task1}
For executing the task, the process begins with the assembly of the manipulator according to the desired morphology and adjusting it to the initial joint state. The initial joint state is also the warm-start of the MPC controller, which has record during the optimization as we introduced in the SEC.~\ref{subsec:warmstart}'s third paragraph. 
%The most efficient initial configuration state and its performance evaluation are recorded to optimize the morphology and mounted pose. These warm-start findings are also valuable for application in real-world experimental conditions.
Subsequently, MPC is applied as the controller for the modular manipulator with the given warm-start to track the reference trajectory in the task space.  
The experimental execution of the task, which considers different optimization objectives, can be found in the attached video. Notably, the mobile platform pose represents the initial pose corresponding to the desired mounted pose of the base link. At this juncture, the end-effector's pose of the manipulator corresponds to the first desired point on the trajectory.
During the execution of the task the effort of the robot joints remained within the maximum continues limits considered. It also shows that the robot avoids collisions with the mobile base of the robot and the wall in the environment. 
%with surrounding environments.

% To further illustrate the practicality in the experiment, we marked the desired drilling positions on a wall with blue dots. A camera mounted on the robot's end effector was then used to detect these dots, allowing for the generation of accurate task space trajectories. After completing a drilling task, the robot proceeds to the vicinity of the next marked point. Then, generating a new drilling trajectory based on the updated camera data. Utilizing the camera's input, we were able to reduce the discrepancy between the actual drilling positions and those anticipated by the simulation, enhancing the experiment's accuracy.

\section{Conclusion}
This paper introduces a computational framework for simultaneously optimizing modular manipulators' physical structure and mounted pose, ensuring their customization for specific tasks. Initially, we developed a mapping function that converts the discrete morphology state into a continuous modular state. This approach enables the \textit{implicit} representation of the morphology state, allowing it to be optimized concurrently alongside the mounted pose of the manipulator within continuous space. 
% Consequently, the optimization challenge has been reformulated into a black-box optimization problem. 
Compared to our previous work, which tackled such issues through combinatorial optimization, this methodology accelerates the convergence speed and facilitates the attainment of more optimal solutions. Finally, we conducted experiments using the CONCERT modular mobile manipulation robot in practical scenarios, such as executing drilling tasks, a common application in construction environments.  

\section*{Acknowledge}
This paper was supported by the Horizon 2020 Research and Innovation
Program of the European Union under Project CONCERT (Grant agreement no 101016007).

\bibliographystyle{ieeetr}
\bibliography{reference.bib}

\end{document}